\documentclass{article} 
\usepackage{iclr2027_conference,times}

\usepackage{amsmath,amsfonts,bm}

\def\eqref#1{equation~\ref{#1}}

\def\1{\bm{1}}

\DeclareMathAlphabet{\mathsfit}{\encodingdefault}{\sfdefault}{m}{sl}
\SetMathAlphabet{\mathsfit}{bold}{\encodingdefault}{\sfdefault}{bx}{n}

\usepackage{hyperref}
\usepackage{url}
\usepackage{booktabs}
\usepackage{multirow} 
\usepackage{graphicx} 
\usepackage{xspace} 
\usepackage{subfig}
\usepackage{caption}

\newcommand{\sysname}{\textit{WUSPruner}\xspace}

\title{WUSPruner: Second-Moment-Optimal Transforms for Structured Pruning}

\iclrfinalcopy
\author{Shengkun Tang \\
MBZUAI\\
\And
Dan Alistarh \\
ISTA \\
\And
Zhiqiang Shen \\
MBZUAI\\
}

\begin{document}

\maketitle
\lhead{Preprint} 
\begin{abstract}

Structured pruning is typically formulated as identifying which model dimensions to remove, often within the model's original representation. However, the effect of coordinate deletion fundamentally depends on the basis in which pruning is performed. This raises a more fundamental question: \emph{how should the basis be chosen specifically for coordinate deletion?} We introduce \sysname, a transform-based framework for structured pruning of large language models. We formulate basis selection as a local linear reconstruction problem that directly minimizes the reconstruction loss induced by pruning. Inspired by the second-moment transformation in WUSH~\citep{chen2026wush}, we consider a WUS transform constructed jointly from activation and weight statistics. We show that under positive-definite second moments, WUS transform minimizes the local reconstruction loss over all invertible changes of basis. In the resulting basis, the activation and weight second moments are simultaneously diagonalized and balanced, causing the deletion error to decompose into independent spectral contributions and yielding a principled criterion for coordinate selection. Guided by this result, we develop a scalable pruning method based on blockwise WUS transformations and compact input projections. Experiments on Llama and Qwen models from 8B to 72B parameters show consistent improvements over structured pruning baselines across language modeling, knowledge, commonsense reasoning, mathematics, and code generation benchmarks, while also delivering practical reductions in inference latency and memory usage.

\end{abstract}
\section{Introduction}

Large language models achieve strong performance at substantial computational and memory cost, motivating compression techniques that reduce the cost of inference. Among them, structured pruning~\citep{sun2024simple, ashkboos2024slicegpt, tang2026slimqwen,tang2025darwinlm} is particularly attractive because it removes entire dimensions or channels, yielding smaller dense operators that can translate into practical reductions in latency and memory usage regardless of the hardware specification. However, structured pruning is also inherently destructive: removing a representation dimension eliminates its contribution to the computation rather than merely zeroing individual parameters without constraint. As a result, even moderate pruning can substantially distort the outputs of the original model. The central challenge is therefore to remove model dimensions while preserving the performance of the original dense model as much as possible. In this work, we study this problem through the local reconstruction of linear mappings, where the intermediate dimension is reduced while minimizing the resulting output distortion.

A key observation is that structured pruning is inherently basis-dependent. Many existing pruning heuristics operate directly in the model's native representation, assigning importance scores to weights or dimensions and removing those deemed least important. Methods such as Wanda~\citep{sun2024simple} perform pruning in the model's original representation, whereas transform-based approaches such as SliceGPT~\citep{ashkboos2024slicegpt} modify the basis before deletion. This highlights that pruning quality depends not only on coordinate importance, but also on the choice of representation basis. 
However, the PCA-based transformation of SliceGPT is chosen according to activation statistics rather than the reconstruction error induced by pruning itself.  This leaves a fundamental question open: \emph{how should the basis be chosen specifically for coordinate deletion?} 
Our goal is therefore to identify a transformation that minimizes the reconstruction loss induced by coordinate deletion under a fixed pruning budget.

In this work, we formalize this objective as a local linear reconstruction problem. Given calibration activations and the corresponding weight matrix, we optimize over invertible changes of basis to minimize the reconstruction loss after deleting a fixed number of transformed coordinates. We show that this objective admits a theoretically optimal solution given by a WUS transformation inspired by the second-moment construction of WUSH~\citep{chen2026wush}. The transform is constructed jointly from activation and weight statistics and is obtained by removing the Hadamard component used for quantization. Under positive-definite activation and weight second moments, WUS attains the minimum reconstruction loss over all invertible transformations. In the WUS basis, the transformed activation and weight second moments are simultaneously diagonalized and balanced, eliminating cross-coordinate interactions in the deletion error. As a result, the reconstruction loss decomposes into independent coordinate-wise contributions determined by a singular-value spectrum, providing a principled criterion for coordinate selection. This characterization further connects transformed coordinate deletion to optimal rank-constrained approximation.

The theoretically optimal construction, however, is not directly suitable for large-scale Transformer pruning. Applying a full transformation to high-dimensional representations incurs substantial construction and storage costs, and practical deployment requires the transformed representation to integrate efficiently with existing Transformer modules. To bridge this gap, we develop \sysname, a scalable approximation to the unrestricted WUS construction. We partition each input space into blocks and compute the WUS transformation locally within each block, substantially reducing the cost of constructing the pruning basis. The retained transformed coordinates are then implemented through compact input projections followed by smaller dense linear maps. Together, these designs make the theoretically motivated basis construction practical for large-scale Transformer pruning while keeping the additional memory and computational overhead manageable.

Empirically, \sysname consistently improves structured pruning quality across model families and scales. We evaluate models ranging from 8B to 72B parameters on language modeling, commonsense reasoning, knowledge, mathematics, and code generation benchmarks. Across the evaluated models, \sysname achieves higher average downstream accuracy than the structured pruning baselines and consistently improves perplexity on PTB and C4. The gains persist when scaling from 8B models to Qwen3.8-27B and Qwen2.5-72B, indicating that the proposed basis construction remains effective at larger model scales. Beyond pruning quality, \sysname also translates structural compression into practical efficiency gains,  including both inference latency and memory consumption reduction. In summary, our main contributions are:

\begin{itemize}
    \item We formulate structured pruning as a \emph{basis-selection problem}: instead of pruning coordinates only in the model's original representation, we optimize the change of basis with respect to the reconstruction loss induced by coordinate deletion.

    \item We provide a theoretical characterization of the optimal basis for this objective. Under positive-definite activation and weight second moments, we show that the WUS transformation minimizes the local reconstruction loss over all invertible transformations. In this basis, the deletion error decomposes into independent spectral contributions, connecting coordinate pruning to optimal rank-constrained approximation.

    \item Building on this result, we develop \sysname, a scalable structured pruning method
    based on blockwise WUS transformations and compact shared input projections. Experiments on Llama and Qwen models from 8B to 72B parameters show consistent improvements over structured pruning baselines across language modeling and downstream tasks, together with practical reductions in inference latency and memory usage.
\end{itemize}
\section{Related Work}
\label{sec:related-work}

\noindent \textbf{Structured and transform-based pruning.}
Post-training pruning methods typically use calibration data or task-derived signals to estimate parameter importance and remove redundant model components. SparseGPT~\citep{frantar2023sparsegpt} enables one-shot unstructured and semi-structured sparsification through approximate second-order reconstruction, while Wanda~\citep{sun2024simple} combines weight magnitudes with input activation norms to construct data-aware pruning scores. Structured pruning instead removes entire coupled components to obtain hardware-friendly compression. LLM-Pruner~\citep{ma2023llmpruner} identifies dependent structures using gradient-based importance and restores performance through low-rank adaptation, ZipLM~\citep{kurtic2023ziplm} explicitly optimizes the trade-off between predictive loss and runtime on a target inference platform, and FLAP~\citep{an2024flap} combines activation fluctuation statistics with adaptive structure selection and bias compensation. A complementary line of work studies whether pruning becomes easier after changing the representation basis. SliceGPT~\citep{ashkboos2024slicegpt}, for instance, exploits orthogonal invariances in Transformers to rotate hidden representations using activation-derived PCA bases and subsequently remove low-importance dimensions. Our proposed \sysname derives the basis from a different objective: {\em minimizing the local output reconstruction error induced by coordinate deletion while jointly accounting for the second moments of both activations and weights.} To translate this principle into a practical structured compressor, our \sysname employs local blockwise projections and reduced linear maps while retaining the original residual-stream dimensionality, and explicitly includes the projection operators in the compressed-model cost.

\noindent \textbf{Activation-aware low-rank compression.}
Low-rank compression~\citep{hu2021lora,dettmers2023qlora,chavan2023one,zhao2024galore,hayou2024lora+,tian2024hydralora} approximates a dense linear operator by the product of smaller matrices, reducing both parameter count and computation. Classical SVD provides the optimal approximation under an unweighted Frobenius objective, whereas recent LLM compression methods incorporate activation statistics to better reflect the actual output distortion induced by compression. ASVD~\citep{yuan2023asvd} rescales weight matrices according to activation distributions before decomposition, mitigating the effect of activation outliers and adapting compression to layer sensitivity.  SVD-LLM~\citep{wang2025svdllm}, for instance, introduces truncation-aware data whitening to connect singular-value truncation with calibration-set reconstruction loss and further performs sequential updates of compressed weights. This viewpoint is closely related to our formulation. After an appropriate change of basis, retaining only $r$ coordinates produces an operator of rank at most $r$. Under the unrestricted local formulation considered in this work, the optimal coordinate-deletion solution achieves the same minimum output reconstruction error as the best activation-weighted rank-$r$ approximation. This yields a spectral characterization that connects structured coordinate pruning with activation-aware low-rank compression. Our \sysname translates the unrestricted theoretical construction into a practical blockwise representation with shared input projections and spectral coordinate selection.

\section{Method}
\label{sec:method}

\begin{figure*}[t]
    \centering
    \includegraphics[width=1\linewidth]{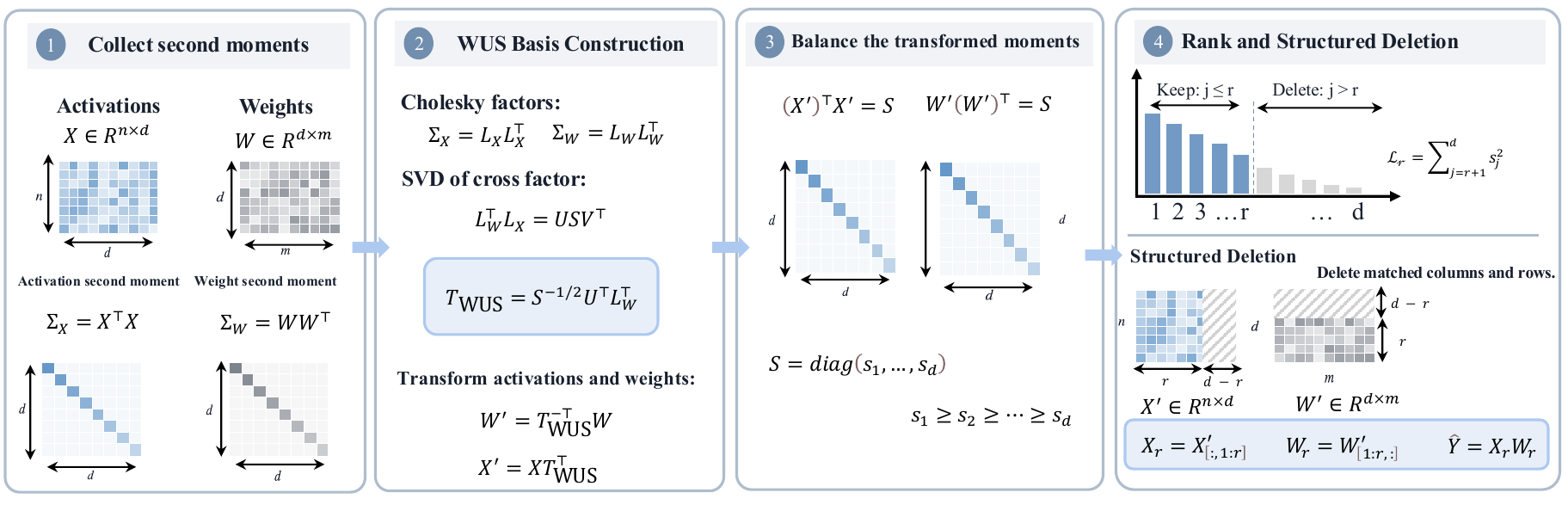}
    \caption{Overview of \sysname. WUS constructs a basis from activation and weight second moments such that the transformed moments are balanced and diagonalized with a shared spectrum $S$. This makes the pruning loss decompose into independent coordinate-wise contributions, allowing transformed coordinates to be ranked by $s_i$ and the least important matched dimensions to be structurally removed.}\
    \label{fig:main_fig}
\vspace{-2em}
\end{figure*}

We introduce \sysname, a transform-based framework for structured pruning.
We first formulate pruning as coordinate deletion under an invertible change of
basis and characterize the reconstruction error induced by deletion. We then
show that the WUS transformation attains the minimum
local reconstruction error over unrestricted invertible transformations under
positive-definite second moments, yielding a spectral criterion for coordinate
selection. Finally, we develop a scalable blockwise approximation for
Transformer LLMs. The overview of \sysname is shown in Figure~\ref{fig:main_fig}.

\subsection{Transformed Coordinate Deletion}
\label{sec:coordinate-deletion}

Consider a linear map $Y=XW$, where
$X\in\mathbb{R}^{n\times d}$ contains calibration activations and
$W\in\mathbb{R}^{d\times m}$ is the weight matrix. Our objective is to reduce
the intermediate dimension from $d$ to $r=d-k$ while preserving the output of
the original map. For any invertible transformation $T\in\mathrm{GL}(d)$, the activations and
weights can be reparameterized without changing the original computation:
$XW=(XT^\top)(T^{-\top}W)$. Let
$P\in\mathbb{R}^{d\times d}$ be a diagonal projector onto the $k$ coordinates
to be deleted. After coordinate deletion, the resulting approximation is
$\widehat{Y}(T)=XT^\top(I-P)T^{-\top}W$.

Although the transformation leaves the computation unchanged before pruning,
the approximation after deletion depends on the chosen basis. We therefore
consider the local reconstruction objective
\begin{equation}
\mathcal{L}(T)
=
\|Y-\widehat{Y}(T)\|_F^2
=
\|XT^\top P T^{-\top}W\|_F^2.
\label{eq:deletion-objective}
\end{equation}
Without loss of generality, we take $P$ to select the last $k$ coordinates,
since any other selection can be represented by a permutation of the
transformed basis.

Define the unnormalized second moments
$\Sigma_X=X^\top X$ and $\Sigma_W=WW^\top$. By cyclic invariance of the trace, ~\eqref{eq:deletion-objective} becomes
\begin{equation}
\mathcal{L}(T)
=
\operatorname{Tr}\!\left(
P\,M_X(T)\,P\,M_W(T)
\right),
\label{eq:deletion-trace}
\end{equation}
where
$M_X(T)=T\Sigma_XT^\top$ and
$M_W(T)=T^{-\top}\Sigma_WT^{-1}$.
Thus, under this local objective, the effect of coordinate deletion depends
jointly on the transformed activation and weight second moments. We do not
center the activations, since the objective concerns reconstruction of the
original linear outputs; a shared bias, if present, cancels from the
reconstruction error.

To make the role of the basis explicit, let
$\widetilde X=XT^\top$ and $\widetilde W=T^{-\top}W$.
For a deleted coordinate set $\mathcal D$, the removed contribution is
$Y-\widehat Y=\sum_{i\in\mathcal D}
\widetilde X_{:,i}\widetilde W_{i,:}$.
Its squared reconstruction error can be written as
\begin{equation}
\begin{aligned}
\mathcal L(T;\mathcal D)
&=
\sum_{i\in\mathcal D}
[M_X(T)]_{ii}[M_W(T)]_{ii} +
2\sum_{\substack{i,j\in\mathcal D\\i<j}}
[M_X(T)]_{ij}[M_W(T)]_{ij}.
\end{aligned}
\label{eq:deletion-cross-terms}
\end{equation}
The first term represents individual coordinate contributions, while the
second captures interactions between deleted coordinates. A basis that removes
these interactions turns coordinate selection into independent reconstruction
costs.

\subsection{Optimal Local Deletion with WUS}
\label{sec:wus-optimality}

We next characterize a transformation that minimizes
~\eqref{eq:deletion-objective}. Assume
$\Sigma_X\succ0$ and $\Sigma_W\succ0$, with Cholesky factorizations
$\Sigma_X=L_XL_X^\top$ and $\Sigma_W=L_WL_W^\top$.
Consider the singular value decomposition
$L_W^\top L_X=USV^\top$, where
$S=\operatorname{diag}(s_1,\ldots,s_d)$ and
$s_1\geq\cdots\geq s_d>0$.

Inspired by the second-moment transformation in WUSH~\citep{chen2026wush}, we define the WUS transformation:
\begin{equation}
T_{\mathrm{WUS}}
=
S^{-1/2}U^\top L_W^\top.
\label{eq:wus-transform}
\end{equation}

Its key property is that it simultaneously diagonalizes and balances the two
second moments:
\begin{equation}
M_X(T_{\mathrm{WUS}})
=
M_W(T_{\mathrm{WUS}})
=
S.
\label{eq:balanced-moments}
\end{equation}
Indeed, $T_{\mathrm{WUS}}^{-\top}W
=S^{1/2}U^\top L_W^{-1}W$, and
$L_W^{-1}\Sigma_WL_W^{-\top}=I$ gives the transformed weight second moment
$S$. On the activation side,
$L_W^\top\Sigma_XL_W=US^2U^\top$, which yields the same transformed second
moment $S$.

Therefore, all off-diagonal interaction terms in~\eqref{eq:deletion-cross-terms} vanish. For any deleted coordinate
set $\mathcal D$,
\begin{equation}
\mathcal{L}(T_{\mathrm{WUS}};\mathcal D)
=
\sum_{i\in\mathcal D}s_i^2.
\label{eq:wus-coordinate-error}
\end{equation}
Each transformed coordinate consequently has an explicit deletion cost, and
retaining coordinates associated with the largest singular values minimizes
the error within the WUS basis.

\medskip
\noindent\textbf{Theorem 1 (Optimal Local Coordinate Deletion).}
\textit{
Suppose $\Sigma_X\succ0$ and $\Sigma_W\succ0$. Let $P$ delete $k$
coordinates and construct $T_{\mathrm{WUS}}$ as in
~\eqref{eq:wus-transform}. Then
}
\begin{equation}
\min_{T\in\mathrm{GL}(d)}\mathcal{L}(T)
=
\mathcal{L}(T_{\mathrm{WUS}})
=
\sum_{i=d-k+1}^{d}s_i^2.
\label{eq:optimal-deletion-error}
\end{equation}

\noindent\textit{Proof.}
For any admissible $T$, the approximation $\widehat{Y}(T)$ has rank at most
$r=d-k$. By the Eckart--Young--Mirsky theorem,
$\mathcal{L}(T)\geq\sum_{i=r+1}^{d}\sigma_i(XW)^2$.
Let $Z=XL_X^{-\top}$, so that $Z^\top Z=I$ and
$XW=ZL_X^\top W$. Hence, $XW$ and $L_X^\top W$ have the same nonzero singular
values. Moreover,
$(L_X^\top W)(L_X^\top W)^\top
=(L_W^\top L_X)^\top(L_W^\top L_X)$,
and therefore $\sigma_i(XW)=s_i$.
~\eqref{eq:balanced-moments} shows that WUS attains this lower bound.
\hfill$\square$

\medskip

The theorem establishes optimality for a fixed linear map and fixed
calibration activations over unrestricted invertible transformations. It does
not imply optimality for the full nonlinear network. WUS is also not the
unique minimizer: diagonal rescaling of its transformed coordinates leaves the
retained and deleted contributions unchanged. The result instead
characterizes an optimal deletion construction and establishes a direct
connection to rank-constrained approximation: WUS realizes the best rank-$r$
local output reconstruction error through a change of basis followed by
coordinate deletion.

\subsection{Practical Design for Transformer LLMs}
\label{sec:blockwise-wus}

The unrestricted construction above provides a theoretical solution to the
local deletion problem, but applying a full transformation to high-dimensional
Transformer representations is expensive. We therefore develop a scalable
approximation based on local input projections, shared transformations, and
blockwise WUS construction.

\noindent \textbf{Integration into attention and MLP modules.}
We replace each selected linear map with an input projection followed by a
smaller dense linear map while preserving its output width. The projection is
applied immediately before the linear map, allowing the compressed module to
retain the original interfaces with normalization, nonlinearities, and
residual connections. In particular, the residual-stream width and input
embeddings remain unchanged. Each transformation is constructed from the activations entering its
associated linear map and the corresponding weights. Maps operating on
different intermediate representations use separate transformations, whereas
maps sharing the same input can share both the basis and projected
activations. The same construction can also be applied to the language-model
output head while preserving the vocabulary dimension.

\noindent \textbf{Sharing projections across linear maps.}
For linear maps $W^{(1)},\ldots,W^{(J)}$ sharing the same input, we construct a common basis from the concatenated weight matrix
$W_{\mathrm{group}}=[W^{(1)}\;\cdots\;W^{(J)}]$. Its weight second moment is 
$\Sigma_{W,\mathrm{group}}
=\sum_{j=1}^{J}W^{(j)}W^{(j)\top}$.
The resulting reconstruction objective is the sum of the individual map
errors, allowing the same input projection to be shared across the group. We use this construction for the gate and up projections. Each map retains its own compressed weight, while the shared projection is stored and evaluated only once. This sharing amortizes the additional parameter cost introduced by the transformation across multiple linear maps, reducing the storage overhead of the compressed representation.

\noindent \textbf{Scalable basis construction.}
To avoid constructing a full $d\times d$ transformation, we partition the input coordinates into disjoint blocks
$\{\mathcal B_b\}_{b=1}^{B}$. For each block, let
$X_b=X_{:,\mathcal B_b}$ and $W_b=W_{\mathcal B_b,:}$, with second moments
$\Sigma_{X,b}=X_b^\top X_b$ and $\Sigma_{W,b}=W_bW_b^\top$.
We then compute an independent WUS transformation $T_b$ within each block using~\eqref{eq:wus-transform}. For block width $h$, the activation moments require $O(dh)$ storage instead of $O(d^2)$, while the Cholesky and SVD operations act on matrices of size at most $h\times h$, resulting in an aggregate factorization cost of
$O(dh^2)$ for equal-sized blocks. If a moment matrix is not positive definite, we add diagonal damping until its Cholesky factorization succeeds. Larger blocks more closely approximate the unrestricted construction but require greater preprocessing and storage cost. The blockwise formulation therefore provides a practical trade-off between approximation quality and construction overhead. Let
$s_{b,1}\geq\cdots\geq s_{b,|\mathcal B_b|}$
denote the WUS spectrum within block $b$. From~\eqref{eq:wus-coordinate-error}, coordinate $i$ has local deletion
cost $s_{b,i}^2$. Given a total retention budget $r$, we allocate
$r_b$ retained coordinates to each block approximately in proportion to its
size, with $\sum_b r_b=r$, and retain the $r_b$ coordinates with the largest
spectral contributions. Thus, coordinate selection requires only sorting
within each block and does not involve gradient-based optimization.

\noindent \textbf{Compact execution and storage.}
Let $R_b\in\mathbb{R}^{r_b\times|\mathcal B_b|}$ select the retained
coordinates in block $b$. The compressed computation is
\begin{equation}
\widehat Y
=
\sum_{b=1}^{B}
\underbrace{X_bT_b^\top R_b^\top}_{\text{projected activations}}
\underbrace{R_bT_b^{-\top}W_b}_{\text{compressed weights}}.
\label{eq:compressed-computation}
\end{equation}
Defining the compact input projection $C_b=R_bT_b$ and compressed weight
$\overline W_b=R_bT_b^{-\top}W_b$, each block contribution becomes
$(X_bC_b^\top)\overline W_b$. The projected activations from all blocks are
concatenated and consumed by a smaller dense linear map, reducing its input
dimension while preserving its output dimension. If all coordinates are
retained, the original computation is recovered exactly because
$T_b^\top T_b^{-\top}=I$.

Ignoring biases, a compressed map stores
$rm+\sum_b r_b|\mathcal B_b|$ scalar values compared with $dm$ in the
original map. For equal block width $h$, this becomes $rm+rh$. The projection
storage is therefore included in the reported compression budget. For a group
of maps sharing the same projection, this cost is counted only once.
\section{Experiments}
\label{sec:exp}
\subsection{Experimental Setup}

\noindent \textbf{Models and Datasets.} We use the WikiText~\citep{merity2016wikitext} training set as the calibration data for most experiments, sampling 128 sequences with a sequence length of 2048. For the mathematics and code generation experiments, we instead use a mixed calibration set of 256 samples, consisting of 128 samples from \texttt{theblackcat102/evol-codealpaca-v1}~\citep{luo2023wizardcoder} and 128 samples from \texttt{allenai/tulu-3-sft-personas-math}~\citep{lambert2024tulu}. All baseline methods use the same calibration data and calibration configuration as \sysname to ensure a fair comparison. We evaluate our method across multiple model families, including Llama3-8B~\citep{grattafiori2024llama}, Qwen3-8B~\citep{yang2025qwen3}, and Qwen3.5-9B~\citep{qwen35blog}. To further evaluate scalability, we extend the experiments to Qwen3.8-27B and Qwen2.5-72B.

\noindent \textbf{Baselines.}
We compare \sysname with three representative compression baselines:
Wanda-sp~\citep{sun2024simple}, SliceGPT~\citep{ashkboos2024slicegpt}, and SVD-LLM~\citep{wang2025svdllm}.
Wanda-sp is a structured variant of Wanda that ranks removable dimensions using weight magnitude and activation statistics.
SliceGPT applies activation-based PCA rotations before dimension deletion.
SVD-LLM performs activation-aware low-rank compression using truncation-aware whitening and singular value decomposition.
For all comparisons, baseline methods use the same calibration data and calibration configuration as \sysname.

\noindent \textbf{Evaluation.}
We evaluate the pruned models from three complementary perspectives.
For language modeling, we report perplexity on WikiText-2~\citep{merity2016wikitext}, PTB~\citep{marcus-etal-1993-ptb}, and C4~\citep{raffel2020c4}.
For commonsense reasoning and knowledge, we report accuracy on PIQA~\citep{Bisk2020}, HellaSwag~\citep{zellers2019hellaswag}, ARC-Challenge~\citep{allenai:arc}, MMLU~\citep{hendryckstest2021mmlu}, OpenBookQA~\citep{OpenBookQA2018}, and CMMLU~\citep{li2024cmmlu}.
We further evaluate mathematical reasoning and code generation on GSM-8K~\citep{cobbe2021gsm8k}, GSM-Platinum, HumanEval~\citep{chen2021humaneval}, HumanEval-Plus, MBPP~\citep{austin2021program}, and MBPP-Plus.
Unless otherwise specified, all downstream evaluations are conducted using the \textit{lm-evaluation-harness} package~\citep{eval-harness}, and HumanEval and MBPP are evaluated using the EvalPlus framework~\citep{evalperf}.

\noindent \textbf{Implementation Details.}
We implement \sysname in PyTorch with Hugging Face Transformers and compress decoder layers sequentially, using activations propagated through previously pruned layers for calibration. WUS transformations are constructed from activation and weight second moments using Cholesky factorization followed by SVD; diagonal damping is applied when needed for numerical stability. For Llama, Qwen2/Qwen2.5, and Qwen3, we prune the attention query and output projections together with the MLP gate, up, and down projections, sharing the input transformation between the gate and up projections. For Qwen3.5, we use the same targets in full-attention layers and prune only the MLP projections in linear-attention layers. We additionally compress the language-modeling head along its input dimension while preserving the vocabulary output. Latency is measured on a single L40S GPU with batch size 1 and sequence length 2048. We use 3 warm-up runs and average over 10 measured runs with 64 decoding steps. All experiments are conducted on 10$\times$L40S GPU server.

\subsection{Main Results.}
\begin{table*}[t]
  \centering
  \caption{Main results on commonsense reasoning, knowledge, and language modeling benchmarks across three LLMs with 25\% compression ratio. We report zero-shot accuracy (\%) and perplexity (PPL), where higher accuracy and lower perplexity indicate better performance. Bold values denote the best results among pruning methods, excluding the dense model.}
  \vspace{-1em}
  \resizebox{1\textwidth}{!}{
  \begin{tabular}{l|c|ccccccc|cccc}
    \toprule
    \multirow{2}{*}{\textbf{Model}} &\multirow{2}{*}{\textbf{Method}}  &\multicolumn{7}{c}{\textbf{Zero-shot Acc.} $\uparrow$}  &\multicolumn{4}{|c}{\textbf{Perplexity} $\downarrow$} \\
      & &PIQA &HellaSwag &ARC-C &MMLU &OBQA &CMMLU &Avg. &Wiki2 &PTB &C4 &Avg. \\
    \midrule
    \multirow{4}{*}{Llama3-8B}
    & Dense &80.3 &79.2 &53.6 &61.9 &45.0 &48.5 &61.4 &6.1 &9.9 &9.3 &8.4 \\
    & Wanda-sp &65.2 &34.5 &28.8 &26.5 &29.8 &25.1 &34.9 &253.6 &336.7 &313.7 &301.3 \\
    & SliceGPT &62.8 &49.6 &31.4 &25.0 &\textbf{34.0} &25.3 &38.0 &\textbf{12.1} &116.1 &121.7 &83.3 \\
    & \sysname (Ours) &\textbf{69.2} &\textbf{53.8} &\textbf{32.5} &\textbf{37.2} &33.0 &\textbf{29.4} &\textbf{42.5} &14.8 &\textbf{24.8} &\textbf{31.3} &\textbf{23.6} \\
    \hline
    \multirow{4}{*}{Qwen3-8B}
    & Dense &77.8 &74.7 &56.5 &72.9 &42.6 &76.9 &66.9 &9.7 &15.4 &15.2 &13.4 \\
    & Wanda-sp &65.8 &46.4 &28.6 &38.1 &29.8 &29.9 &39.7 &46.0 &39.2 &43.3 &42.8 \\
    & SliceGPT &67.6 &\textbf{55.5} &35.3 &31.6 &33.4 &28.6 &42.0 &\textbf{13.1} &67.8 &43.8 &41.5 \\
    & \sysname (Ours) &\textbf{67.8} &53.7 &\textbf{39.2} &\textbf{54.3} &\textbf{34.6} &\textbf{47.5} &\textbf{49.5} &15.0 &\textbf{28.6} &\textbf{33.7} &\textbf{25.7} \\
    \hline
    \multirow{4}{*}{Qwen3.5-9B}
    & Dense &80.3 &78.0 &55.8 &78.6 &42.8 &80.3 &69.3 &8.6 &12.9 &11.9 &11.1 \\
    & Wanda-sp &67.9 &33.6 &\textbf{48.3} &36.0 &31.0 &33.3 &41.6 &53.5 &62.6 &44.3 &53.4 \\
    & SliceGPT &66.2 &55.3 &37.0 &28.5 &34.2 &25.3 &41.0 &\textbf{11.2} &36.7 &33.7 &27.2 \\
    & \sysname (Ours) &\textbf{69.0} &\textbf{56.0} &39.0 &\textbf{54.5} &\textbf{34.4} &\textbf{53.8} &\textbf{51.1} &12.4 &\textbf{20.7} &\textbf{21.8} &\textbf{18.3} \\
    \bottomrule
  \end{tabular}}
  \label{tab:main}
  \vspace{-1em}
\end{table*}

\begin{table*}[t]
  \centering
  \caption{Performance on mathematical reasoning and code generation benchmarks using Qwen3.5-9B with 25\% compression ratio. We report accuracy (\%) across six benchmarks and their average.}
  \vspace{-1em}
  \resizebox{0.9\textwidth}{!}{
  \begin{tabular}{l|ccccccc}
    \toprule
    \multirow{2}{*}{\textbf{Method}}
    & \multicolumn{7}{c}{\textbf{Zero-shot Acc.} $\uparrow$} \\
    & HumanEval & HumanEval-Plus & MBPP & MBPP-Plus & GSM-8K & GSM-Platinum & Avg. \\
    \midrule

    Dense
    & 84.8 & 79.9 & 78.6 & 63.8 & 87.4 & 70.6 & 77.5 \\

    Wanda-sp
    & 4.9 & 4.3 & 11.1 & 9.5 & 4.2 & 3.1 & 6.2 \\

    SliceGPT
    & 62.8 & 55.5 & 54.2 & 44.1 & 56.4 & 19.3 & 48.7 \\

    \sysname (Ours)
    & \textbf{64.0} & \textbf{57.3} & \textbf{57.1}
    & \textbf{49.5} & \textbf{58.9} & \textbf{37.3}
    & \textbf{54.0} \\

    \bottomrule
  \end{tabular}}
  \label{tab:coding}
\end{table*}

\noindent \textbf{Results on Commonsense Benchmarks.} We compare \sysname with Wanda-sp and SliceGPT across three language models, as summarized in Table~\ref{tab:main}. \sysname consistently achieves the highest average accuracy among the pruning methods, obtaining 42.5\%, 49.5\%, and 51.1\% on Llama3-8B, Qwen3-8B, and Qwen3.5-9B, respectively. These results correspond to improvements of 4.5, 7.5, and 9.5 percentage points over the strongest baseline for each model. Moreover, \sysname achieves the highest accuracy on five out of six individual benchmarks for each model, demonstrating its effectiveness across different model architectures and evaluation tasks. Notably, \sysname exhibits substantial improvements on knowledge-intensive benchmarks. On MMLU, it achieves accuracy gains of 10.7, 16.2, and 18.5 percentage points over the strongest baseline for Llama3-8B, Qwen3-8B, and Qwen3.5-9B, respectively. Similar improvements are observed on CMMLU, where \sysname outperforms the strongest baseline by 4.1, 17.6, and 20.5 percentage points. These results indicate that \sysname is particularly effective at preserving knowledge-related capabilities under structured pruning. 

In terms of language modeling performance, \sysname achieves the lowest average perplexity across all three models, reducing it from 83.3 to 23.6 on Llama3-8B, from 41.5 to 25.7 on Qwen3-8B, and from 27.2 to 18.3 on Qwen3.5-9B compared with the strongest baseline in each case. Interestingly, while SliceGPT achieves slightly lower perplexity on WikiText-2, its performance deteriorates substantially on PTB and C4, suggesting a potential sensitivity to the calibration data distribution. Since WikiText is used for calibration, this discrepancy may indicate that SliceGPT's activation-based transformation is more closely adapted to the calibration distribution, potentially limiting its generalization to other text distributions. In contrast, \sysname maintains consistently lower perplexity on PTB and C4 across all three models, suggesting improved robustness across evaluation datasets. These findings highlight the importance of jointly considering activation and weight statistics when constructing pruning transformations to preserve language modeling performance beyond the calibration distribution.
Together, these results demonstrate that \sysname provides a more favorable balance between downstream task accuracy and language modeling quality across the evaluated models.

\noindent \textbf{Results on Math and Coding Benchmarks.} Table ~\ref{tab:coding} evaluates the mathematical reasoning and coding capabilities of the pruned Qwen3.5-9B model across six benchmarks. \sysname consistently outperforms both Wanda-sp and SliceGPT on all evaluated tasks, achieving an average accuracy of 54.0\%, compared with 48.7\% for SliceGPT and 6.2\% for Wanda-sp. In particular, \sysname achieves substantial improvements on GSM-Platinum, increasing accuracy from 19.3\% to 37.3\%, an absolute gain of 18.0 percentage points over SliceGPT. It also achieves improvements of 5.4 and 2.5 percentage points on MBPP-Plus and GSM-8K, respectively. These results suggest that \sysname better preserves mathematical reasoning and coding capabilities under structured pruning, particularly on challenging tasks that require multi-step reasoning and precise code generation. The results on commonsense benchmarks can be found in the Appendix~\ref{app:math_coding}.

\noindent \textbf{Scaling Up to Larger Models.} To evaluate the scalability of \sysname, we extend our experiments to Qwen3.8-27B and Qwen2.5-72B, as shown in Table~\ref{tab:larger_model}. \sysname consistently outperforms Wanda-sp and SliceGPT on both models, achieving average accuracies of 60.9\% and 63.9\%, respectively. These results represent improvements of 9.0 and 7.5 percentage points over the strongest baseline for each model. Notably, \sysname achieves substantial gains on knowledge-intensive benchmarks, outperforming the strongest baseline by 13.7 and 11.7 percentage points on MMLU, and by 16.6 and 16.6 percentage points on CMMLU for the 27B and 72B models, respectively. In terms of language modeling performance, \sysname achieves the lowest average perplexity on both models, reducing it from 21.3 to 13.1 on Qwen3.8-27B and from 18.2 to 13.1 on Qwen2.5-72B compared with the strongest baseline. Consistent with the observations on smaller models, \sysname achieves lower perplexity on PTB and C4, while SliceGPT performs better on WikiText-2. These results suggest that \sysname maintains its effectiveness across different model scales, preserving both downstream task accuracy and language modeling quality in larger language models.

\begin{table*}[t]
  \centering
  \caption{Evaluation of \sysname on larger language models, including Qwen3.8-27B and Qwen2.5-72B with 25\% compression ratio.}
  \vspace{-1em}
  \resizebox{1\textwidth}{!}{
  \begin{tabular}{l|c|ccccccc|cccc}
    \toprule
    \multirow{2}{*}{\textbf{Model}} &\multirow{2}{*}{\textbf{Method}}  &\multicolumn{7}{c}{\textbf{Zero-shot Acc.} $\uparrow$}  &\multicolumn{4}{|c}{\textbf{Perplexity} $\downarrow$} \\
      & &PIQA &HellaSwag  &ARC-C &MMLU &OBQA &CMMLU &Avg.  &Wiki2 &PTB & C4 &Avg. \\
    \midrule
    \multirow{4}{*}{Qwen3.8-27B}& Dense &81.6   &82.9  &58.5  &83.4  &46.2    &71.7  &70.7   &7.0   &8.8   &11.0  &8.9\\
    \
    &Wanda-sp &71.6   &66.1  &37.3  &55.4  &37.2  &43.9  &51.9  &24.5  &24.2  &25.7  &24.8\\
    &SliceGPT &70.1   &63.7  &43.0  &37.2  &40.2  &28.6  &47.1  &\textbf{8.7}  &29.3  &25.9  &21.3\\
    &\sysname (Ours) &\textbf{75.7} &\textbf{68.6} &\textbf{51.1} &\textbf{69.1} &\textbf{40.6} &\textbf{60.5} &\textbf{60.9} &9.1 &\textbf{13.3} &\textbf{16.9} &\textbf{13.1}\\
    \hline
    \multirow{4}{*}{Qwen2.5-72B}& Dense &83.6   &86.0  &62.7  &83.3  &46.2 &89.9  &75.2   &3.8   &8.4   &9.1  &7.1 \\
    \
    &Wanda-sp &70.1   &65.4  &47.1  &59.3  &\textbf{43.4} &53.2  &56.4  &12.9  &24.5  &\textbf{17.2}  &18.2\\
    &SliceGPT &73.2   &69.2  &43.3  &60.3  &40.6  &49.6  &56.0  &\textbf{6.4}  &48.0  &26.1  &26.8\\
    &\sysname (Ours) &\textbf{76.2} &\textbf{71.2} &\textbf{53.0} &\textbf{72.0} &41.4 &\textbf{69.8} &\textbf{63.9} &8.1 &\textbf{13.9} &17.5 &\textbf{13.1}\\
    \bottomrule
  \end{tabular}}
  \vspace{-1em}
  \label{tab:larger_model}
\end{table*}

\subsection{Analysis}
\begin{figure*}[t]
\begin{center}
\subfloat[Different Block Sizes]{\includegraphics[width=0.33\textwidth]{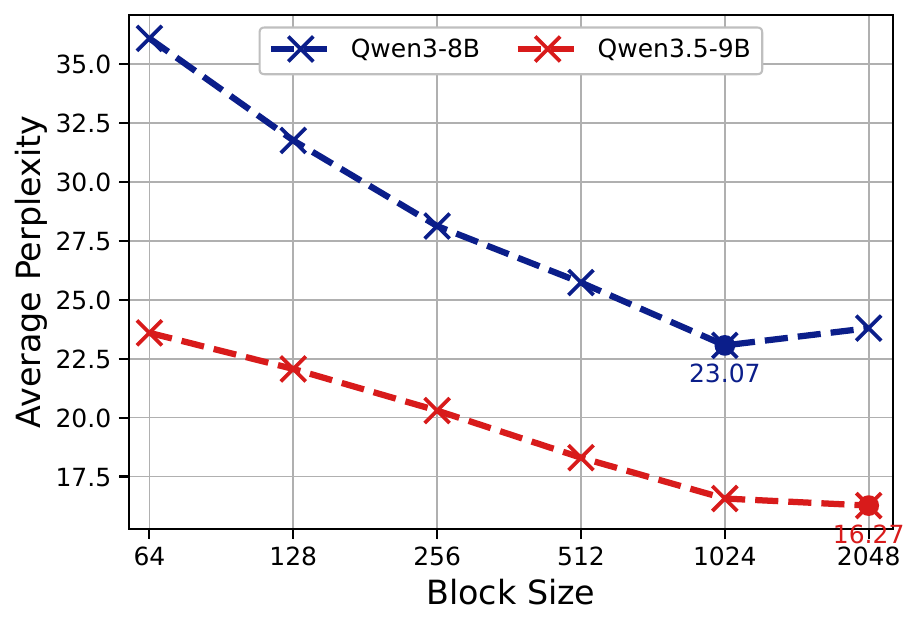}}
\subfloat[Different Calibration Sizes]{\includegraphics[width=0.33\textwidth]{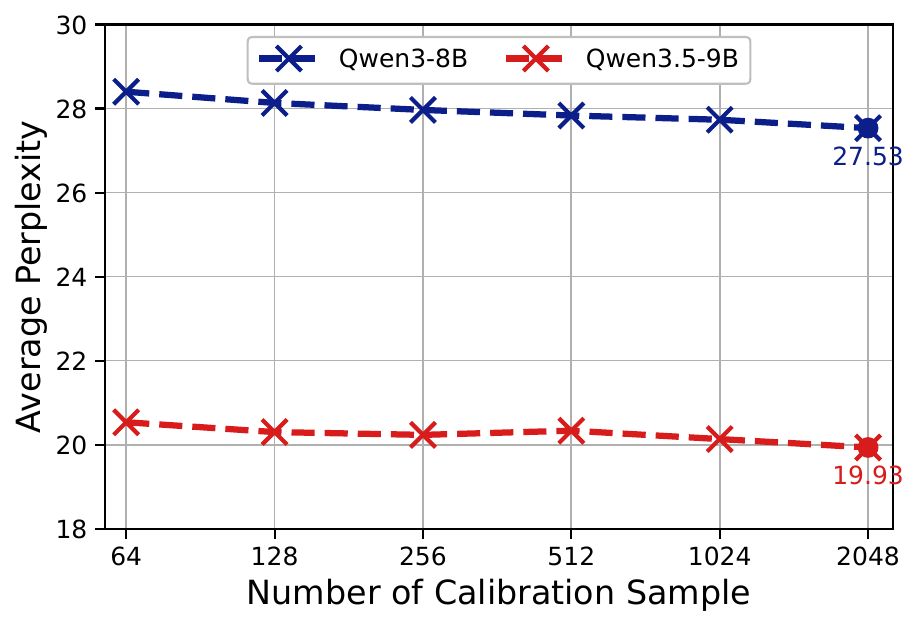}}
\subfloat[Different Sparsity]{\includegraphics[width=0.33\textwidth]{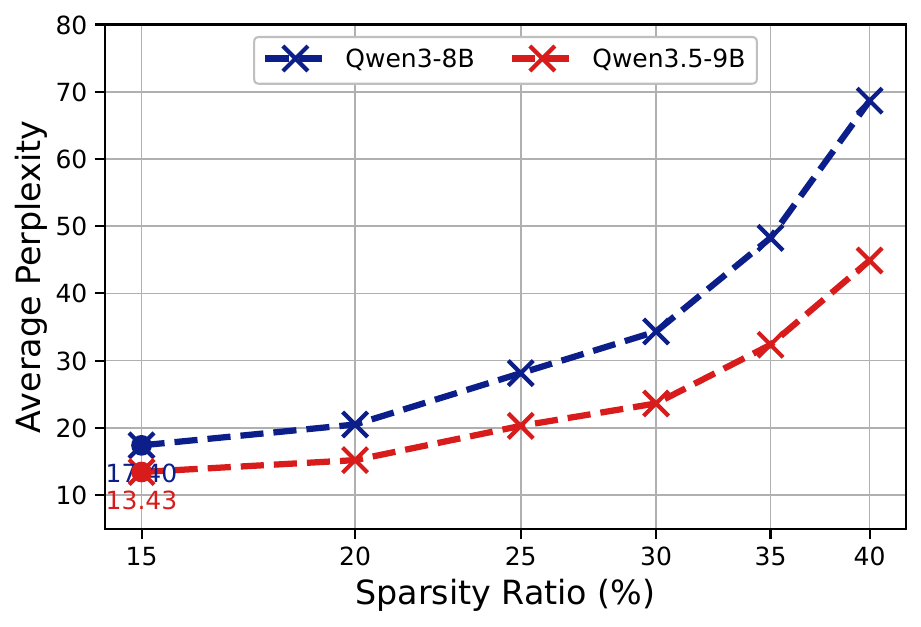}}
\end{center}
\vspace{-1em}
\caption{Sensitivity analysis of WUSPruner with respect to block size, calibration set size, and pruning ratio on Qwen3-8B and Qwen3.5-9B. Larger block sizes generally yield lower perplexity, consistent with a closer approximation to the original WUS transformation, while increasing computational overhead. WUSPruner remains stable across a wide range of calibration set sizes, and its performance degrades smoothly as the pruning ratio increases.}
\label{Fig:sensitivity}
\vspace{-.5em}
\end{figure*}

\noindent \textbf{Sensitivity to Block Size, Calibration Size, and Sparsity.} We evaluate the sensitivity of \sysname to block size, calibration-set size, and pruning sparsity, as shown in Figure~\ref{Fig:sensitivity}. We first observe that increasing the block size consistently improves performance on both Qwen3-8B and Qwen3.5-9B, as reflected by the decreasing average perplexity. This behavior is consistent with our formulation: a larger block relaxes the blockwise constraint and more closely approximates the original WUS transformation derived in Section~\ref{sec:wus-optimality}. However, larger blocks also incur higher computational and memory costs when constructing the transformation, leading to a practical trade-off between pruning quality and preprocessing overhead. In contrast, \sysname is largely insensitive to the number of calibration samples. Increasing the calibration set from 64 to 2048 samples results in only minor variations in perplexity for both models, indicating that the proposed transformation can be estimated reliably from a relatively small calibration set. Finally, as expected, performance gradually degrades as the pruning ratio increases, with the degradation becoming more pronounced under aggressive sparsity. Nevertheless, both models exhibit consistent trends across sparsity levels, suggesting that \sysname behaves stably over a broad range of compression ratios. Overall, these results show that \sysname benefits from larger transformation blocks while remaining robust to calibration-set size and exhibiting predictable degradation as compression becomes more aggressive. Additional results across compression ratios are provided in Appendix~\ref{app:compression_ratio}.

\noindent \textbf{Comparison with SVD-based Compression.} We compare \sysname with SVD-LLM on Qwen3-8B at a 25\% sparsity ratio, using the same calibration and evaluation setup, as shown in Table~\ref{tab:svd_ppl}. \sysname consistently outperforms SVD-LLM across all three datasets. While the two methods achieve comparable perplexity on WikiText-2 (15.0 vs. 15.1), \sysname substantially reduces perplexity on PTB from 45.9 to 28.6 and on C4 from 93.3 to 33.7, leading to a much lower average perplexity of 25.7 compared with 51.4 for SVD-LLM. These results suggest that \sysname preserves language modeling quality more effectively across different data distributions under the same compression ratio. Additional benchmark comparisons are provided in Appendix~\ref{app:svd}.
\begin{table*}[t]
    \centering

    \begin{minipage}[t]{0.38\textwidth}
        \centering
        \scriptsize
        \setlength{\tabcolsep}{4pt}
        \captionof{table}{
            Perplexity evaluation of \sysname compared with SVD-LLM at a 25\% sparsity ratio.
        }
        \vspace{-1em}
        \renewcommand{\arraystretch}{1.0}

        \resizebox{\linewidth}{!}{%
        \begin{tabular}{l|cccc}
            \toprule
            \multirow{2}{*}{\textbf{Method}}
            & \multicolumn{4}{c}{\textbf{Perplexity} $\downarrow$} \\
            & Wiki2 & PTB & C4 & Avg. \\
            \midrule

            Dense & 9.7 & 15.4 & 15.2 & 13.4 \\
            SVD-LLM &15.1  &45.9  &93.3  &51.4   \\
            \sysname (Ours) & \textbf{15.0} & \textbf{28.6} & \textbf{33.7} & \textbf{25.7} \\

            \bottomrule
        \end{tabular}%
        }
        
        \label{tab:svd_ppl}
    \end{minipage}
    \hfill
    \begin{minipage}[t]{0.6\textwidth}
        \centering
        \scriptsize
        \setlength{\tabcolsep}{4pt}
        \captionof{table}{
            Speedup and memory reduction. Speedup and peak GPU memory
            reduction are measured relative to the dense model under the
            same inference configuration.
        }
        \vspace{-1em}
        \renewcommand{\arraystretch}{1.0}

        \resizebox{\linewidth}{!}{%
        \begin{tabular}{c|c|cc|cc}
            \toprule
            \multirow{2}{*}{\textbf{Sparsity}} &
            \multirow{2}{*}{\textbf{Method}} &
            \multicolumn{2}{c|}{\textbf{Inference Efficiency}} &
            \multicolumn{2}{c}{\textbf{GPU Memory}} \\
            & &
            Latency (ms) $\downarrow$ &
            Speedup $\uparrow$ &
            Peak (GB) $\downarrow$ &
            Reduction $\uparrow$ \\
            \midrule

            - & Dense
            & 183.9 & $1.00\times$ & 16.2 & - \\

            25\% & \sysname
            & 160.8 & $1.14\times$ & 13.2 & -18.5\% \\

            50\% & \sysname
            & 124.1 & $1.48\times$ & 9.4 & -42.0\% \\

            \bottomrule
        \end{tabular}%
        }

        \label{tab:speedup_memory}
    \end{minipage}
\vspace{-1em}
\end{table*}

\noindent \textbf{Speedup and Memory Reduction.}
Table~\ref{tab:speedup_memory} reports the practical inference benefits of \sysname under different pruning ratios. Compared with the dense model, \sysname consistently reduces both inference latency and peak GPU memory usage, with larger gains obtained at higher sparsity. At 25\% sparsity, the latency decreases from 183.9 ms to 160.8 ms, while peak GPU memory is reduced from 16.2 GB to 13.2 GB. When the sparsity is increased to 50\%, the latency further decreases to 124.1 ms and the peak memory usage drops to 9.4 GB. This demonstrates that the structural compression introduced by \sysname translates into tangible runtime benefits, rather than merely reducing the nominal parameter count. Moreover, the improvements become more pronounced as the pruning ratio increases, highlighting the potential of \sysname for reducing both computational and memory costs in resource-constrained deployment scenarios. Additional runtime results are provided in Appendix~\ref{app:runtime}.






\section{Conclusion}
We presented \sysname, a transform-based framework for structured pruning of large language models. By formulating pruning as coordinate deletion under a change of basis, we show that the WUS transformation minimizes local reconstruction error over invertible transformations and yields a principled spectral criterion for coordinate selection. We further develop a practical blockwise WUS approximation with compact shared projections for Transformer LLMs. Experiments on models from 8B to 72B demonstrate consistent improvements over structured pruning baselines, together with practical reductions in inference latency and memory usage. These results highlight basis optimization as a principled and effective direction for structured LLM compression.
\section*{AI use statement}

In this work, we used generative AI tools for assist in the writing of proofs, design or provide feedback on research  methodology or experiments, implement methods.
We have not used generative AI tools for assist with translation, support qualitative and thematic data analysis, interpret results, propose or refine hypotheses,
and Generate synthetic data sets, help develop theoretical models or conceptual frameworks, formulate mathematical claims, provide critical ingredients for proving mathematical claims, clean and reformat dataset, are not applicable to this work.
Additionally, we used generative AI tools for create or edit software code. We have reviewed all AI-assisted work. We checked the LLM-generated code carefully and tested all results multiple times. We take responsibility for the final content of this work,
including text, claims or artifacts produced with the aid of generative AI.

\section*{Ethical Statement}
This work aims to improve the efficiency of LLM deployment by reducing memory and inference costs. Positive impacts include improved accessibility of capable language models in resource-constrained environments and reduced computational cost. Potential negative impacts include making capable generative models easier to deploy for misuse, such as spam or misinformation generation. The method does not introduce new model capabilities, and standard safeguards for the underlying LLMs should still be applied after compression.

\section*{Reproducibility statement}
We describe the formulation and algorithm of \sysname in Section~\ref{sec:method}, and provide the experimental settings and implementation details in Section~\ref{sec:exp}. These descriptions and analyses clarify the conditions under which our results are obtained and support their reproduction.

\bibliography{iclr2027_conference}
\bibliographystyle{iclr2027_conference}

\appendix
\newpage
\appendix

\section{Notation Summary}

\begin{table}[h]
\centering
\caption{Summary of the main notation used in the paper.}
\label{tab:notation}
\small
\resizebox{0.9\textwidth}{!}{
\begin{tabular}{c|l|c}
\toprule
\textbf{Symbol} & \textbf{Description} & \textbf{Shape / Domain} \\
\midrule

$X$
& Calibration activations
& $\mathbb{R}^{n\times d}$ \\

$W$
& Weight matrix of a linear layer
& $\mathbb{R}^{d\times m}$ \\

$d$
& Original input dimension
& -- \\

$r$
& Number of retained coordinates
& -- \\

$T$
& Invertible change-of-basis transformation
& $\mathrm{GL}(d)$ \\

$P$
& Diagonal projector selecting deleted coordinates
& $\mathbb{R}^{d\times d}$ \\

$\mathcal{L}(T)$
& Local output reconstruction loss after coordinate deletion
& $\mathbb{R}_{\geq 0}$ \\

\midrule

$\Sigma_X=X^\top X$
& Activation second-moment matrix
& $\mathbb{R}^{d\times d}$ \\

$\Sigma_W=WW^\top$
& Weight second-moment matrix
& $\mathbb{R}^{d\times d}$ \\

$L_X,L_W$
& Cholesky factors of $\Sigma_X$ and $\Sigma_W$
& $\mathbb{R}^{d\times d}$ \\

$S=\mathrm{diag}(s_1,\ldots,s_d)$
& WUS spectrum obtained from $L_W^\top L_X=USV^\top$
& $\mathbb{R}^{d\times d}$ \\

$T_{\mathrm{WUS}}$
& WUS transformation, $S^{-1/2}U^\top L_W^\top$
& $\mathbb{R}^{d\times d}$ \\

$s_i^2$
& Reconstruction cost associated with deleting coordinate $i$
& $\mathbb{R}_{>0}$ \\

\midrule

$B_b$
& Coordinate set of block $b$
& $B_b\subseteq\{1,\ldots,d\}$ \\

$X_b,\;W_b$
& Activation and weight submatrices for block $b$
& -- \\

$T_b$
& Blockwise WUS transformation
& $\mathbb{R}^{|B_b|\times |B_b|}$ \\

$r_b$
& Number of retained coordinates in block $b$
& -- \\

$R_b$
& Selector of retained coordinates in block $b$
& $\mathbb{R}^{r_b\times |B_b|}$ \\

$C_b=R_bT_b$
& Compact input projection
& $\mathbb{R}^{r_b\times |B_b|}$ \\

$\overline{W}_b$
& Compressed weight matrix
& $\mathbb{R}^{r_b\times m}$ \\

\bottomrule
\end{tabular}}
\end{table}

\section{More Experimental Results}
\subsection{More Benchmark Results of Model Calibrated by Math and Coding Data.}
\label{app:math_coding}
\begin{table}[h]
  \centering
  \caption{Additional evaluation of Qwen3.5-9B calibrated with the mixed mathematics and coding dataset at a 25\% compression ratio.}
  \resizebox{1\textwidth}{!}{
  \begin{tabular}{l|ccccccc|cccc}
    \toprule
    \multirow{2}{*}{\textbf{Method}}
    & \multicolumn{7}{c}{\textbf{Zero-shot Acc.} $\uparrow$}
    & \multicolumn{4}{|c}{\textbf{Perplexity} $\downarrow$} \\
    & PIQA & HellaSwag & ARC-C & MMLU & OBQA & CMMLU & Avg.
    & Wiki2 & PTB & C4 & Avg. \\
    \midrule

    Dense
    & 80.3 & 78.0 & 55.8 & 78.6 & 42.8 & 80.3 & 69.3
    & 8.6 & 12.9 & 11.9 & 11.1 \\

    Wanda-sp
    & 67.4 & 43.5 & 33.1 & 45.0 & \textbf{31.0} & 41.0 & 43.5
    & 73.5 & 74.7 & 67.1 & 71.8 \\

    SliceGPT
    & 58.3 & 37.8 & 27.2 & 32.5 & 27.4 & 29.0 & 35.4
    & 53.1 & 58.6 & 57.6 & 56.4 \\

    \sysname (Ours)
    & \textbf{68.1} & \textbf{50.5} & \textbf{38.9} & \textbf{52.1}
    & 30.6 & \textbf{54.1} & \textbf{49.1}
    & \textbf{21.8} & \textbf{29.8} & \textbf{28.9} & \textbf{26.8} \\

    \bottomrule
  \end{tabular}}
  \label{tab:math_coding_calibration_general}
\end{table}

To examine whether the choice of calibration domain affects general-purpose capabilities, we further evaluate models calibrated with the mixed mathematics and coding dataset on commonsense, knowledge, and language modeling benchmarks. \sysname remains consistently stronger than the pruning baselines, achieving the highest accuracy on five of the six downstream benchmarks and improving the average accuracy to 49.1\%, compared with 43.5\% for Wanda-sp and 35.4\% for SliceGPT. The advantage is particularly pronounced in language modeling, where \sysname achieves an average perplexity of 26.8, substantially lower than 71.8 for Wanda-sp and 56.4 for SliceGPT. These results indicate that \sysname is relatively robust to the calibration domain and better preserves general model capabilities even when the calibration data are specialized toward mathematics and coding.

\subsection{More Benchmark Comparison with SVD-LLM.}
\label{app:svd}
\begin{table}[h]
  \centering
  \caption{Additional comparison with SVD-LLM on Qwen3-8B at a 25\% compression ratio.}
  \resizebox{1\textwidth}{!}{
  \begin{tabular}{l|ccccccc|cccc}
    \toprule
    \multirow{2}{*}{\textbf{Method}}
    & \multicolumn{7}{c}{\textbf{Performance} $\uparrow$}
    & \multicolumn{4}{|c}{\textbf{Perplexity} $\downarrow$} \\
    & PIQA & HellaSwag & ARC-C & MMLU & OBQA & CMMLU & Avg.
    & Wiki2 & PTB & C4 & Avg. \\
    \midrule

    Dense
    & 77.8 & 74.7 & 56.5 & 72.9 & 42.6 & 76.9 & 66.9
    & 9.7 & 15.4 & 15.2 & 13.4 \\

    SVD-LLM
    & 67.6 & \textbf{54.3} & 38.9 & 45.5 & \textbf{38.2} & 37.7 & 47.0
    & 15.1 & 45.9 & 93.3 & 51.4 \\

    \sysname (Ours)
    & \textbf{67.8} & 53.7 & \textbf{39.2} & \textbf{54.3} & 34.6
    & \textbf{47.5} & \textbf{49.5}
    & \textbf{15.0} & \textbf{28.6} & \textbf{33.7} & \textbf{25.8} \\

    \bottomrule
  \end{tabular}}
  \label{tab:svdllm_more_benchmarks}
\end{table}
We further compare \sysname with SVD-LLM on Qwen3-8B at a 25\% compression ratio. Overall, \sysname achieves stronger downstream performance, improving the average accuracy from 47.0\% to 49.5\%. The gains are particularly pronounced on knowledge-intensive benchmarks, with improvements of 8.8 and 9.8 percentage points on MMLU and CMMLU, respectively, while SVD-LLM performs better on HellaSwag and OpenBookQA. \sysname also substantially improves language modeling quality, reducing the average perplexity from 51.4 to 25.8, with especially large reductions on PTB and C4. These results show that the proposed transform-based pruning approach provides a better overall trade-off between downstream task performance and language modeling quality than activation-aware low-rank compression under the same compression ratio.

\subsection{Benchmark Performance with Different Compression Ratio}
\label{app:compression_ratio}

\begin{table*}[h]
  \centering
  \caption{
  Benchmark performance of \sysname under different compression ratios on Qwen3-8B and Qwen3.5-9B.
  }
  \resizebox{\textwidth}{!}{
  \begin{tabular}{l|c|ccccccc|cccc}
    \toprule
    \multirow{2}{*}{\textbf{Model}}
    & \multirow{2}{*}{\textbf{Sparsity}}
    & \multicolumn{7}{c}{\textbf{Zero-shot Acc.} $\uparrow$}
    & \multicolumn{4}{|c}{\textbf{Perplexity} $\downarrow$} \\
    &
    & PIQA & HellaSwag & ARC-C & MMLU & OBQA & CMMLU & Avg.
    & Wiki2 & PTB & C4 & Avg. \\
    \midrule

    \multirow{6}{*}{Qwen3-8B}
    & 15\% & 74.1 &64.7  &50.6  & 64.8 & 39.6& 64.1 & 59.5 &12.0 &19.3  &20.9 &17.4  \\
    & 20\% &71.2  &59.4  &43.5  &60.2  &37.6  &56.3  &54.7  &13.5  & 22.8 &25.2  &20.5  \\
    & 25\% & 67.8 & 53.7 & 39.2 & 54.3 & 34.6 & 47.5 & 49.5
           & 15.0 & 28.6 & 33.7 & 25.8 \\
    & 30\% &63.0  &47.8  &33.1  &47.7  &30.6  &36.8  &43.1  &17.5  &41.2  &44.3  &34.3  \\
    & 35\% &60.7  &42.3  &28.2  &37.7  &30.6  &29.9  &38.2  &21.4  &59.1  &64.3  &48.2  \\
    & 40\% &59.1  &37.8  &24.4  &30.2  &28.8  &25.7  &34.3  &27.2  &87.4  &91.4  &68.6  \\
    \midrule

    \multirow{6}{*}{Qwen3.5-9B}
    & 15\% &75.9  &68.8  &51.5  &70.6  &40.0  &68.9  &62.6  & 9.8 &15.3  &15.2 & 13.4 \\
    & 20\% &73.8  &63.3  &46.3  &65.0  &37.8  &61.1  &57.8 
    &10.8 & 17.2 &17.6  & 15.2 \\
    & 25\% & 69.0 & 56.0 & 39.0 & 54.5 & 34.4 & 53.8 & 51.1
           & 12.4 & 20.7 & 21.8 & 18.3 \\
    & 30\% &64.4  &48.1  &31.7  &47.5  &32.4  &44.6  &44.7  &14.9  &27.2  &28.8  &23.6  \\
    & 35\% &61.3  &42.3  &27.6  &33.9  &29.0  &35.7  &38.3  &18.4  &39.0  &39.7  &32.3  \\
    & 40\% &58.7  &36.8  &23.7  &26.9  &26.0  &28.2  &33.3  &23.5  &55.1  &56.3  &44.9  \\

    \bottomrule
  \end{tabular}}
  \label{tab:different_compression_ratios}
\end{table*}

\noindent \textbf{Performance across Compression Ratios.}
We further evaluate \sysname on Qwen3.5-9B under compression ratios ranging from 15\% to 40\%. As the compression ratio increases, performance degrades smoothly and consistently across all evaluated benchmarks. The average downstream accuracy decreases from 62.6\% at 15\% compression to 51.1\% at 25\%, while the average perplexity increases moderately from 13.4 to 18.3. More aggressive compression leads to a more pronounced degradation: at 40\%, the average accuracy drops to 33.3\% and the average perplexity rises to 44.9. Similar monotonic trends are observed across individual commonsense, knowledge, and language modeling benchmarks. These results indicate that \sysname provides a predictable trade-off between compression ratio and model quality, with moderate compression preserving substantially more of the original model performance than aggressive compression.

\subsection{Runtime of \sysname.}
\label{app:runtime}
\begin{table}[h]
  \centering
  \caption{
  Runtime comparison of different compression methods.
  We report the total wall-clock time required to compress each model using the same calibration setting.
  }
  \resizebox{0.8\columnwidth}{!}{
  \begin{tabular}{l|cccc}
    \toprule
    \textbf{Method}
    & \textbf{Qwen3-8B}
    & \textbf{Qwen3.5-9B}
    & \textbf{Qwen3.8-27B}
    & \textbf{Qwen2.5-72B} \\
    \midrule

    Wanda-sp
    & 0.2 & 0.3 & 0.8 & 1.1 \\



    \sysname (Ours)
    & 0.5 & 0.67 & 1.5 & 2.3 \\

    \bottomrule
  \end{tabular}}
  \label{tab:runtime}
\end{table}

We further measure the wall-clock time required to compress models of different scales using the same calibration setting as in the main experiments. As expected, \sysname introduces additional preprocessing cost compared with the lightweight Wanda-sp baseline, since it requires collecting second-moment statistics and constructing the WUS transformations. Nevertheless, the overall compression cost remains modest: \sysname requires only 0.5 and 0.67 hours for Qwen3-8B and Qwen3.5-9B, respectively, and 1.5 and 2.3 hours for Qwen3.8-27B and Qwen2.5-72B. The 8B and 9B models are processed on a single GPU, while the 27B and 72B models use 4 and 6 GPUs, respectively, primarily to accommodate the larger models in memory. Overall, the additional preprocessing introduced by WUS is small relative to model deployment and remains practical even at the 72B scale.

\end{document}